\pdfoutput=1

\documentclass[11pt]{article}

\usepackage[]{EMNLP2022}

\usepackage{times}
\usepackage{latexsym}

\usepackage{graphicx}
\usepackage{amsmath}
\usepackage{amssymb}
\usepackage{booktabs}

\graphicspath{ {./images/} }
\usepackage{subcaption}
\usepackage{float}
\usepackage{sidecap}
\usepackage{multirow}

\usepackage[T1]{fontenc}

\usepackage[utf8]{inputenc}

\usepackage{microtype}

\usepackage{inconsolata}

\title{Learning to Model Multimodal Semantic Alignment for Story Visualization}

\author{Bowen Li \\
  University of Oxford \\
  \texttt{bowen.li@cs.ox.ac.uk} \\\And
  Thomas Lukasiewicz \\
  TU Wien,
  University of Oxford \\
  \texttt{thomas.lukasiewicz@cs.ox.ac.uk} \\}

\begin{document}
\maketitle
\begin{abstract}
Story visualization aims to generate a sequence of images to narrate each sentence in a multi-sentence story, where the images should be realistic and keep global consistency across dynamic scenes and characters. Current works face the problem of semantic misalignment because of their fixed architecture and diversity of input modalities. To address this problem, we explore the semantic alignment between text and image representations by learning to match their semantic levels in the GAN-based generative model. More specifically, we introduce dynamic interactions according to learning to dynamically explore various semantic depths and fuse the different-modal information at a matched semantic level, which thus relieves the text-image semantic misalignment problem. Extensive experiments on different datasets demonstrate the improvements of our approach, neither using segmentation masks nor auxiliary captioning networks, on image quality and story consistency, compared with state-of-the-art methods. 
\end{abstract}

\section{Introduction}
\label{sec:intro}
Story visualization is a challenging task, which aims to generate a sequence of story images given a multi-sentence story, and further requires output images to be consistent, e.g., having a consistent background or character appearance. Regardless of its difficulties, story visualization has the potential for many applications, including art creation, computer-aided design, and image editing. 

To address the challenges, current methods~\cite{li2019storygan,song2020character,maharana2021improving,maharana2021integrating} adopt a fixed StoryGAN-based~\cite{li2019storygan} architecture, where two GANs~\cite{goodfellow2014generative} are adopted, one for single-image quality, and one for story consistency, without considering the semantic alignment between different-modal text and image features involved in the generation process.

So, one problem arising is that a fixed network with the involvement of different-modal representations (e.g., text and image) may suffer from a semantic misalignment problem. 
This is because current methods usually adopt a fixed text encoder and image encoder to extract corresponding features, and then use these features directly in an also fixed GAN-based network. However, text representations can be a coarse sentence vector, fine-grained word embeddings, or a structured knowledge graph~\cite{DBLP:conf/icmla/MahonGLL20}, while~\citet{gatys2016image} and \citet{johnson2016perceptual} have shown that image features extracted from different layers of a convolutional neural network (e.g., VGG) may contain different-level semantic information. Based on this, simply fusing cross-domain representations together using a fixed generative network may cause a considerable adverse impact on the quality of output images. For example, one of the main function of the discriminator in a conditional GAN is to evaluate the semantic alignment between input text and output image, and provide the corresponding feedback to the generator, which encourages the generator to generate text-semantic-matched images. However, to evaluate the semantic alignment, the discriminator in current methods only simply concatenates a coarse sentence vector and image features at a small-scale (i.e., 4$\times 4$), extracted from a given image using a series of convolutional layers, which may fail to fully match the semantics between text and image features, and thus provide a less precise training feedback to the generator.

To address this problem, we explore the semantic alignment between text and image representations by learning to match their semantic levels in the GAN-based generative model. 
More specifically, we introduce dynamic interactions according to learning to dynamically explore various semantic depths and fuse the different-modal information at a matched semantic level. By doing this, the network can learn to dynamically utilize various semantic level information from the given representations, and also learn to selectively fuse them together, which thus mitigates the semantic misalignment problem across different modalities. So, the main contributions are summarized as follows:

\noindent $\bullet$  
We fully explore the semantic misalignment problem existing in current methods, and propose a novel single-GAN based network, which improves FID from 78.64 to 52.87, and FSD from 94.53 to 55.20 on Pororo-SV, and establishes a benchmark FID of 74.12 and FSD of 20.07 on Abstract Scenes.

\noindent $\bullet$  We conduct extensive experiments and a thorough analysis of aligning the semantics between different-modal inputs to provide general modeling insights into conditional GANs.

\section{Related Work}
Story visualization aims to generate a sequence of consistent images corresponding to a multi-sentence story. 
StoryGAN~\cite{li2019storygan} introduced a two-GAN-based generation network. CP-CSV~\cite{song2020character}, DUCO~\cite{maharana2021improving}, and VLC~\cite{maharana2021integrating} were built on StoryGAN, where CP-CSV utilized character segmentation masks to improve the performance, and DUCO and VLC adopted auxiliary captioning networks to build a text-image-text circle to ensure the consistency between the input and output. Recently, \citet{li2022word} proposed to utilize fine-grained word information to build a concise single-GAN based network, and~\citet{li2022clustering} proposed to combine both clustering learning and contrastive learning together to ensure better text and image representations in a joint space.
However, all these methods were based on a fixed network without considering the semantic alignment between involved text and image representations. 

Text-to-image generation is closely related to our work, which generates one image from one given text description~\cite{reed2016generative, zhang2018stackgan++,qiao2019mirrorgan, hinz2019semantic, tao2020df, zhu2019dm, xu2018attngan, li2019controllable, li2019object, li2020image, zhang2021cross, li2022memory}. Differently, story visualization is more challenging, as it further requires output story images to be consistent.

\begin{figure*}[t]

\begin{minipage}{1\textwidth}
\includegraphics[width=1\linewidth, height=0.1375\linewidth]{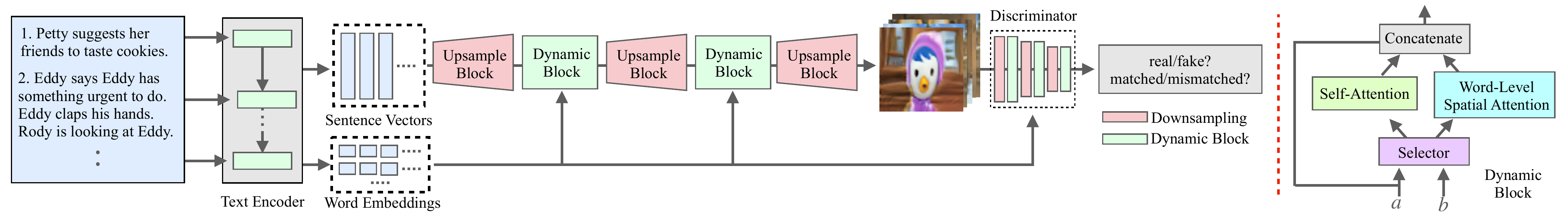}
\end{minipage}

\centering
\vspace{-2ex}
\caption{Architecture of the proposed method (left) and dynamic block (right).}
\label{fig:archi}
\vspace{-2ex}
\end{figure*}

\section{Overview}
Differently from current methods that adopt two GANs, our network only has a single GAN, neither requiring additional segmentation masks nor auxiliary networks for supervision, as we experimentally find that a single-GAN-based network can effectively produce high-quality story images with a good consistency. We attribute this improvement to the exploration of semantic alignment between different-modal representations, which enables fine-grained training feedback from the discriminator to the generator.

Given a story $X$ with $n$ story sentences, a text encoder encodes each story sentence $S_i$ into a sentence vector $s\in \mathbb{R}^{D} $ with corresponding word embeddings $s_\text{word} \in \mathbb{R}^{D \times L}$, where $D$ is the feature dimension, and $L$ is the number of words in a sentence. 
Then, we feed this $n$ sentence vectors into the generation pipeline using upsampling blocks to produce story images at the required resolution. Meanwhile, we further incorporate word embeddings into the generation pipeline, according to our proposed dynamic interactions, allowing the generator to learn to choose semantically aligned inputs from different semantic levels and modalities to achieve a better generation. 
In the discriminator, we also adopt the proposed dynamic interactions to fuse both text and image features, enabling a better evaluation on the text-image semantic alignment. The complete architecture is shown in Fig.~\ref{fig:archi}.

\subsection{Exploration of Semantic Alignment}
To ensure the semantic alignment between text and image representations, we introduce dynamic interactions via utilizing self-attention and cross-modal attention, learning to dynamically explore various semantic depths from these different-modal representations, and also to fuse them together at a matched semantic level, which thus mitigates the text-image semantic misalignment problem.

\subsubsection{Attention Mode}
\label{sec:mode}
Two attention modes are adopted in our approach, one is self-attention~\cite{zhang2019self} (SA), and one is word-level spatial attention~\cite{xu2018attngan} (WSA). To generalize both types of attention, we set $a$ as one input, and $b$ as the other input, where $a$ denotes intermediate image features in the generator or discriminator, and $b$ denotes word embeddings in WSA, or the same image features as $a$ in SA. So, the attention weights can be achieved via $\beta = \text{Softmax}(ab^T)$. Then, we can get weighted hidden features $h$ via $h=\beta b$. Finally, we selectively fuse the weighted hidden features into the network using the proposed dynamic block (details are shown in Section~\ref{sec:attn}). 
Note that both types of attention share the same attention weights, and SA mainly focuses on capturing correlations between long-range pixels within the same image~\cite{li2022lightweight}, and WSA mainly focuses on fusing cross-domain text information with intermediate image features in both the generator and the discriminator at a matched semantic level. 

\subsubsection{Dynamic Block}
\label{sec:attn}
Our proposed dynamic block is in Fig.~\ref{fig:archi}, right. Unlike current methods that only allow the interaction between different modal representations at specific locations with fixed semantics, we propose the dynamic block to learn to achieve a dynamic interaction in an end-to-end manner. Our dynamic block is based on a attention selector, which selectively chooses an appropriate attention (i.e., SA or WSA) to further explore semantic depth or fuse these cross-domain pieces of information together. 

To achieve the selection effect, we first get global representations $\bar{a}$ and $\bar{b}$ of both $a$ and $b$ (see Section~\ref{sec:mode}) using average pooling. Then, the correlation $w$ between $a$ and $b$ can be obtained via $w = \text{Sigmoid}(\bar{a}\bar{b})$, where $w$ denotes the correlation level between $a$ and $b$. 
Then, Gumble-Softmax reparameterization~\cite{jang2016categorical} is adopted to choose a particular attention, based on the probability of each attention, e.g., the probability of SA can be defined as:
\begin{equation}
\small
\vspace{-1ex}
p(\text{SA}) = \frac{\text{exp}((\text{log}(w)+z)/\tau )}{\text{exp}((\text{log}(w)+z)/\tau  + \text{exp}((\text{log}(1-w)+z)/\tau },
\vspace{1ex}
\end{equation}
where $z\,{=}\,-\text{log}(-\text{log}(\mu))$ is sampled Gumble noi\-se, $\mu$ is drawn from the uniform distribution, and $\tau$ is a hyperparameter. Similarly, $p(\text{WSA})$ is obtained from  $p(\text{SA})$ by replacing $w$ with $1-w$ in the numerator.
So, given $a$ and $b$, our dynamic block performs soft weighting in training and hard selection at inference, denoted as:
\vspace{-1ex}
\begin{equation}
h_{\text{soft}} = p(\text{SA})h_{\text{SA}} + p(\text{WSA})h_{\text{WSA}}
\vspace{-4ex}
\end{equation}
\begin{equation}
h_{\text{hard}} = \left\{\begin{matrix}
h_{\text{SA}}, \;\text{if}\; p(\text{SA}) > p(\text{WSA})\\
h_{\text{WSA}}, \;\text{if}\; p(\text{SA}) \leqslant p(\text{WSA}),\\
\end{matrix}\right.
\end{equation}
where $h_\text{SA}$ denotes using self-attention to further explore semantic depth, and $h_\text{WSA}$ denotes fusing finer word information into the generation pipeline.

\subsection{Objective Functions}
The training follows the training procedure of GANs, where the generator and discriminator are trained alternatively by minimizing their losses. 

\section{Experiments}
To evaluate the performance of our approach, we compare it with StoryGAN~\cite{li2019storygan}, CP-CSV~\cite{song2020character}, DUCO~\cite{maharana2021improving}, and VLC~\cite{maharana2021integrating}.

\begin{table*}[t!]
    \centering
    \begingroup
    \setlength{\tabcolsep}{8pt} 
    \caption{Quantitative comparison between different methods on Pororo-SV and Abstract Scenes. For FID, FSD, and the number of parameters, lower is better. For Cosine, higher is better.}
    \label{table:quan}
    \renewcommand{\arraystretch}{1.26} 
    \vspace{-1ex}
    \scalebox{0.9}{
    \begin{tabular}{l|ccc|ccc|cc}
    \hline \hline
    &\multicolumn{3}{c|}{Pororo-SV} &\multicolumn{3}{c|}{Abstract Scenes} &\multicolumn{2}{c}{Number of Trainable Parameters}\\
    Method & FID$\downarrow$ & FSD$\downarrow$ & Cosine$\uparrow$ & FID$\downarrow$ & FSD$\downarrow$ & Cosine$\uparrow$ &  Generator$\downarrow$ & Discriminator$\downarrow$\\
    \hline \hline 
    StoryGAN & 78.64 & 94.53 & 0.22 & 135.16 & 55.80 & 3.59 & 47.0M & 47.2M\\
    CP-CSV & 67.76 & 71.51 & 0.32 & - & - & - & 86.9M & 70.9M  \\
    DUCO & 95.17 & 171.70 & 0.08 & 142.34 & 49.16 & 3.95 & 53.2M & 47.2M  \\
    VLC & 94.30 & 122.07 & 0.21 & - & - & - & 54.5M & 47.2M \\
    \hline
    Ours & \textbf{52.87} & \textbf{55.20} & \textbf{4.61} & \textbf{74.12} & \textbf{20.07} & \textbf{7.28} & \textbf{25.1M} & \textbf{23.5M}\\
    \hline
    \end{tabular}
    }
    \endgroup
    \vspace{-2ex}
\end{table*} 

\subsection{Implementation}
We evaluated our approach at the resolution $64 \times 64$. 
The text encoder is a bi-directional LSTM, pretrained to maximize the cosine similarity between matched image and text features~\cite{xu2018attngan}.
We selected the best checkpoints and tune hyperparameters by using the FID and FSD scores. 
The network was trained for 120 epochs on both Pororo-SV and Abstract Scenes. The Adam optimizer~\cite{kingma2014adam} was adopted with learning rate $0.0002$. We evaluated our approach on a single Quadro RTX 6000 GPU.

\subsection{Datasets}
Pororo-SV is adopted to evaluate our approach, which is built on PororoQA, a dataset for video question answering~\cite{kim2017deepstory}. In Pororo-SV, each story has five consecutive images with corresponding text descriptions. There are $13,000$ story samples in the training set, and $2,336$ story samples in the test set. Differently, we do not evaluate our approach on CLEVR-SV~\cite{li2019storygan}, as there are only $15$ different words in the entire CLEVR-SV dataset, which might fail to fully explore the multimodal network for the story visualization task. We adopt Abstract Scenes~\cite{zitnick2013bringing} to further evaluate our approach. Abstract Scenes was proposed for studying semantic information, which contains over $1,000$ sets of $10$ semantically similar scenes of children playing outside. The scenes are composed of $58$ clip-art objects, and there are six sentences describing different aspects of a scene. In this dataset, we treat scenes from the same set as a story, as they are all created from the same seed scene, sharing similar semantic information. 

\subsection{Evaluation Metrics}
The Fr\'echet inception distance (FID)~\cite{heusel2017gans} and the Fr\'echet story distance (FSD)~\cite{song2020character} are adopted as quantitative evaluation metrics to evaluate the performance of our approach. FID computes the Fr\'echet distance between the distribution of real images and the distribution of fake images. Differently from FID focusing on single image, FSD is proposed for the story visualization task, which takes the sequence of images into account. FSD is built on the principle of FID by using R$(2+1)$~\cite{tran2018closer} as backbone model, where R$(2+1)$ has a flexible sequence length and the strong ability to capture temporal consistency. 

However, as both FID and FSD cannot reflect the semantic alignment between sentences and story images, following~\cite{li2022clustering, li2022word}, we compute the average cosine similarity (Cosine) between pairs of sentence and synthetic image over the testing set, and further scale the value by $100$.  

Besides, we show the number of parameters in the generator and the discriminator for different methods on Pororo-SV to compare the size of different networks.

\subsection{Qualitative Evaluation}
\begin{figure}[t]
\begin{minipage}{0.45\textwidth}
\includegraphics[width=1\linewidth, height=1.27\linewidth]{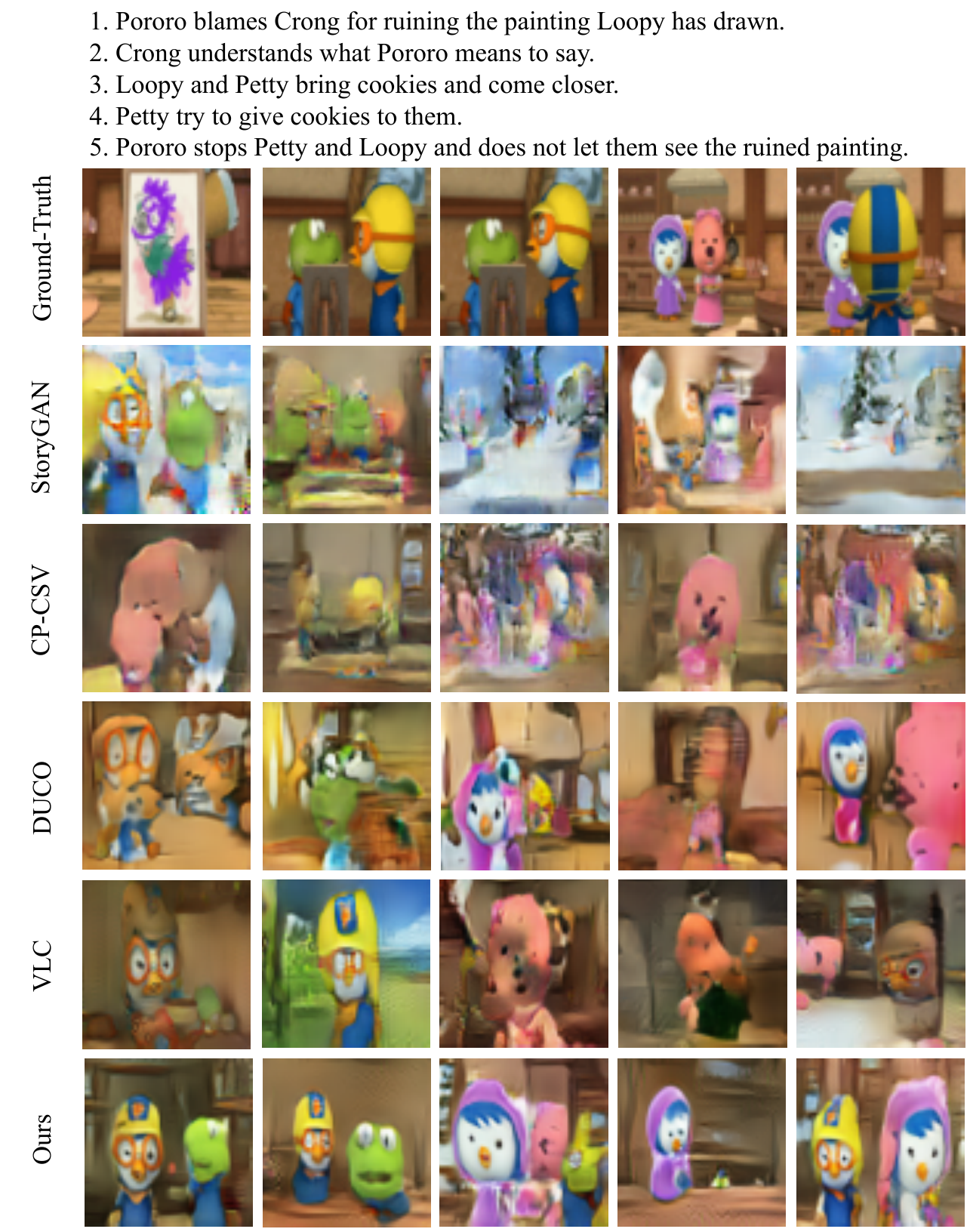}
\end{minipage}

\centering
\vspace{-1ex}
\caption{Qualitative comparison on Pororo-SV.}
\label{fig:qual}
\vspace{-4ex}
\end{figure}

\begin{figure}[t]
\centering
\begin{minipage}{0.42\textwidth}
\includegraphics[width=1\linewidth, height=1.0\linewidth]{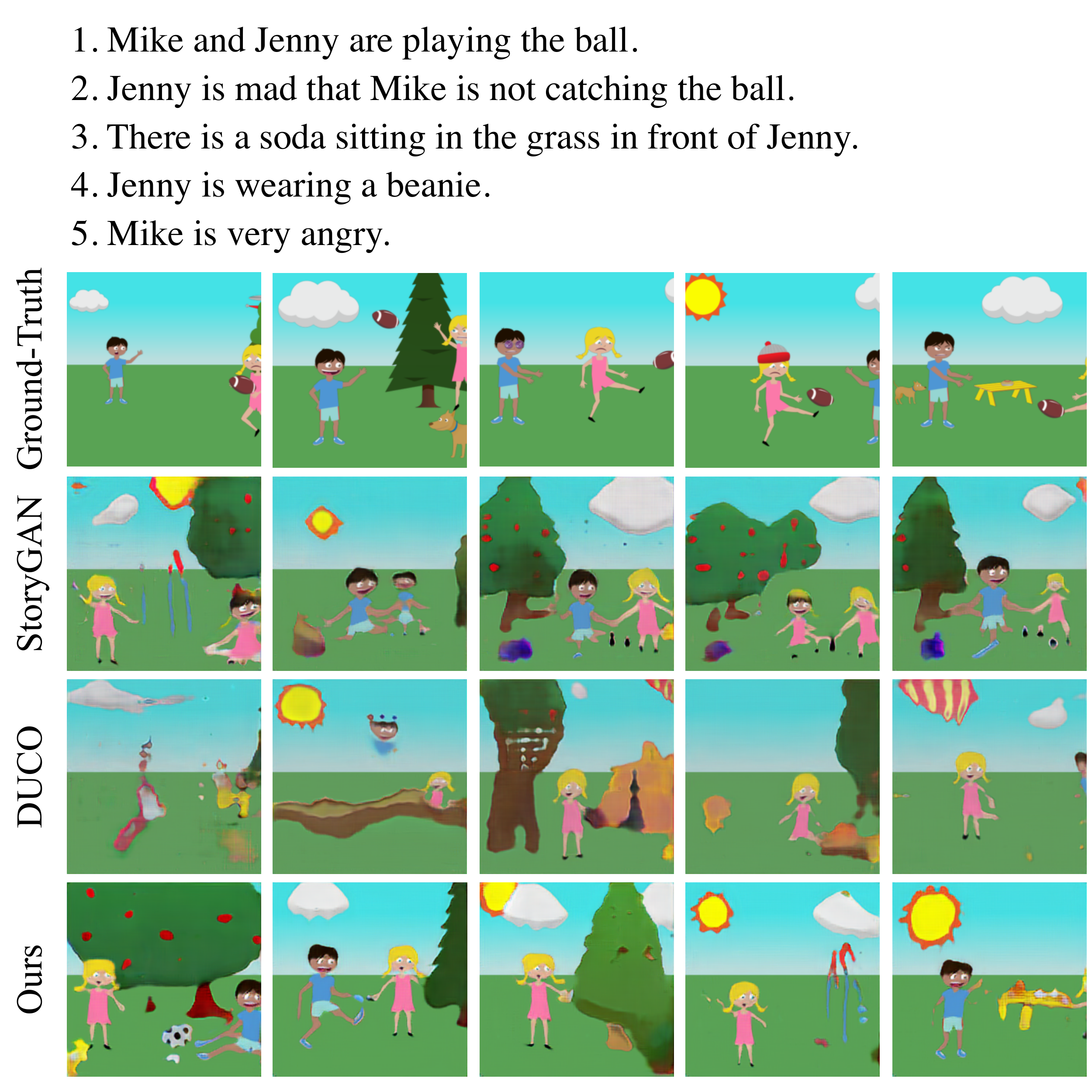}
\end{minipage}

\centering
\vspace{-1ex}
\caption{Qualitative comparison on Abstract Scenes.}
\label{fig:qual2}
\vspace{-2ex}
\end{figure}

Figs.~\ref{fig:qual} and~\ref{fig:qual2} show examples of a visual comparison between our approach and the baselines on Pororo-SV and Abstract Scenes, respectively. Our approach generates realistic images with better regional details, text-image alignment, and consistency, for example, shown in Fig.~\ref{fig:qual}, the characters Pororo (i.e., penguin) and Crong (i.e., frog) have a sharper shape with fine-grained regional details, such as hats and glasses, and shown in Fig.~\ref{fig:qual2}, our approach generates a ball, aligned with the given first sentence, while other methods fail to generate the required ball object on the grass.

\subsection{Quantitative Evaluation}
Table~\ref{table:quan} shows a quantitative comparison with the following widely used metrics on Pororo-SV and Abstract: FID~\cite{heusel2017gans}, FSD~\cite{song2020character}, and Cosine~\cite{li2022clustering}.
From the tables, we can observe that our approach achieves better results against others. 
This illustrates that our approach can generate images with finer quality, achieve better image-text semantic alignment, and keep higher consistency across story images.

We further compare the number of trainable parameters in different methods. As our method is a single-GAN based network, compared to StoryGAN, it reduces the size of the generator by about 46.59\%, and of the discriminator by about 50.21\%.

\subsection{Component Analysis}
Table~\ref{table:ablation} shows a component analysis to evaluate the effectiveness of different components. Without either attention in the dynamic block, the performance of our model degrades, and the worst performance is obtained when the model is without using the entire dynamic block. This demonstrates that (1) each attention mode is important in the dynamic block, and (2) simply fusing different-modal representations without considering their semantics fails to comprehensively improve the performance.

Besides, we further consider the effectiveness of the dynamic block in the generator and discriminator. The worse performance in ``Ours w/o DB in D'' shows that the dynamic block plays a more important role in the discriminator. We think this is because the discriminator needs to provide training feedback to the generator, in terms of image quality and text-image alignment. If the discriminator fails to match the semantics between text and image representations, training feedback may be less precise, which may hinder the generator to generate high-quality story images. The degraded performance in ``Ours w/o DB in G'' is because the generator cannot effectively capture the correlation between text and image representations, even though there is precise and fine-grained training feedback provided by the discriminator. This demonstrates the complementary effect between the adoption of the dynamic block in both the generator and discriminator.

\begin{table}[t]
  \centering
  \caption{Ablation study on Pororo-SV. ``Ours w/o both Attn'' denotes without using both attention in the dynamic block (DB); ``Ours w/ SA Only'' denotes only adopting the self-attention in DB; ``Ours w/ WSA Only'' denotes only adopting the word-level spatial attention in DB; and ``w/o DB in G (or D)'' denotes without using DB in the generator (or discriminator).}
  \vspace*{-1ex}
  \label{table:ablation}
  \smallskip
  \scalebox{0.9}{
  \begin{tabular}{lcccc}
    \toprule
    \multicolumn{1}{c}{Method} &
    \multicolumn{1}{c}{FID} & 
    \multicolumn{1}{c}{FSD} \\
    \midrule
    Ours w/o both Attn & 69.84 & 74.25
     \\
    Ours w/ SA Only & 63.87 & 70.96 \\
    Ours w/ WSA Only & 60.96 & 65.82 \\
    \midrule
    Ours w/o DB in G & 57.10 & 58.93 \\
    Ours w/o DB in D & 61.81 & 60.49 \\
    \midrule
    Ours & 52.87 & 55.20 \\
    \bottomrule
    \vspace*{-3ex}
  \end{tabular}
  }
\end{table}

\subsection{Human Evaluation}
Similarly to~\cite{li2022clustering}, a human evaluation on Pororo-SV is conducted based on three evaluation criteria: (1) visual quality, (2) text-image semantic alignment, and (3) consistency across story images. We asked workers to decide which sample is the best, where each sample contains story images and corresponding sentences. 100 randomly selected samples are assigned to three workers to reduce the human variance. Workers prefer the results that are generated by our approach. 

\begin{table}[t]
  \centering
  \caption{Human evaluation on Pororo-SV between VLC, DUCO, and Ours based on three criteria.}
  \label{table:human_comp}
  \vspace{-1ex}
  \scalebox{0.9}{
  \begin{tabular}{lccc}
    \toprule
    \multicolumn{1}{c}{Choice (\%)} &
    \multicolumn{1}{c}{Ours} & 
    \multicolumn{1}{c}{VLC} & 
    \multicolumn{1}{c}{DUCO} \\
    \midrule
    Visual Quality & 80.33  & 12.00 & 7.67   \\
    \midrule
    Alignment & 77.33 & 13.67 & 9.00  \\
    \midrule
    Consistency  & 82.00 & 9.67 & 8.33 \\
    \bottomrule
  \vspace{-5ex}
  \end{tabular}
  }
\end{table}

\section{Conclusion}
\label{sec:conc}
In this paper, we explored the semantic misalignment problem existing in current story visualization methods, and further proposed dynamic interactions via learning to dynamically explore various semantic depths and fuse the different-modal information at a matched semantic level. 
Experiments demonstrate the superior performance of our proposed single-GAN based approach, with a fewer number of parameters, neither using segmentation masks nor auxiliary captioning networks. 
 
\section*{Acknowledgments}
This work was supported by the Alan Turing Institute under the EPSRC grant EP/N510129/1, by the AXA Research Fund, and by the EPSRC grant EP/R013667/1. We also acknowledge the use of the EPSRC-funded Tier 2 facility JADE (EP/P020275/1) and GPU computing support by Scan Computers International Ltd.

\newpage 
\section*{Limitations}
Our method may fail to generate high-quality results when a given multi-sentence story is complex, e.g., it describes multiple characters (e.g., >\,3) with various backgrounds. Currently, similarly to current methods, our approach focuses more on small-size story image generation, which means that when the number of images in a story is larger (e.g., >\,15), our method may fail to ensure consistency between different story images.

\section*{Ethical Considerations}
The datasets that we use in this paper do not have any personally identifiable information or offensive content, as they are cartoon datasets for educational purposes (Pororo-SV and Abstract Scenes).

\bibliography{main}
\bibliographystyle{acl_natbib}

\end{document}